\documentclass[a4paper, 10pt, conference]{IEEEtran} 

\usepackage[utf8]{inputenc} 
\usepackage[T1]{fontenc}    
\usepackage{hyperref}       
\usepackage{url}            
\usepackage{booktabs}       
\usepackage{amsfonts}       
\usepackage{nicefrac}       
\usepackage{microtype}      
\usepackage{multirow}        
\usepackage{graphicx}        

\newcommand{\RNum}[1]{\uppercase\expandafter{\romannumeral #1\relax}}

\graphicspath{ {images/} }
\title{\LARGE \bf Scalable Object Detection for Stylized Objects}

\author{%
\IEEEauthorblockN{Aayush Garg, Thilo Will, William Darling, Willi Richert, Clemens Marschner} 
\IEEEauthorblockA{Microsoft AI \& Research Munich \\ {\tt\{aayushg,thilow,wdarling,wilrich,clemensm\}@microsoft.com}}
}

\begin{document}

\maketitle

\begin{abstract} Following recent breakthroughs in convolutional neural networks
and monolithic model architectures, state-of-the-art object detection models can
reliably and accurately scale into the realm of up to thousands of classes.
Things quickly break down, however, when scaling into the tens of thousands, or,
eventually, to millions or billions of unique objects. Further, bounding
box-trained end-to-end models require extensive training data. Even though --
with some tricks using hierarchies -- one can sometimes scale up to thousands of
classes, the labor requirements for clean image annotations quickly get out of
control.  In this paper, we present a two-layer object detection method for
brand logos and other stylized objects for which prototypical images exist.  It
can scale to large numbers of unique classes. Our first layer is a
CNN from the Single Shot Multibox Detector family of models that learns to
propose regions where some stylized object is likely to appear. The contents of
a proposed bounding box is then run against an image index that is targeted for
the retrieval task at hand.  
The proposed architecture  scales to a large number of object classes, 
allows to continuously add new classes without retraining, and 
exhibits state-of-the-art quality on a stylized object detection task such as
logo recognition. 
\end{abstract}

\section{Introduction}

Object detection in images is one of the longest running and most important
goals in computer vision. Fast and accurate object detection is a requirement
for numerous AI-dependent applications including self-driving vehicles,
autonomous robotics, assisted photography, and many more. Following recent
breakthroughs in neural networks, and convolutional neural networks in
particular, object detection models have continuously improved over the last
several years. In addition to being fast and accurate, however, object detection
models also need to be in some sense erudite -- or at least as erudite as a
model can be. In short, while object detection with a limited number of output
classes can be useful in several niche applications, the
ultimate goal of object detection is a system that can recognize \textit{all}
objects and where new objects can be continuously added. 

A generic algorithm for this problem is not in sight; the most prominent example 
of scalable specific object detection is face recognition, which operates on the 
constrained space of recognized faces in an image. We propose a solution for 
the domain of \textit{stylized objects}. By this we mean graphical depictions 
like logos which are mainly two-dimensional and for
which prototypical images exist. We use logos as a running example but
the same principles can be applied to movie posters, wine labels, books, etc. 

The idea emerges from the dual-problem of scaling up existing object detection
systems. First, the complexity of training and then serving an object detection
model quickly increases with the number of classes. This problem is similar to
that seen in neural machine translation~\cite{JeanCMB14}, but more
importantly, the quality of the models also quickly breaks down. Even if it were
realistic to be able to obtain the requisite annotation data for millions of  
distinct object classes, differentiating between these classes within an object
detection network has proven to be extremely difficult. 

\begin{figure} 
\centering
\includegraphics[width=1.0\columnwidth]{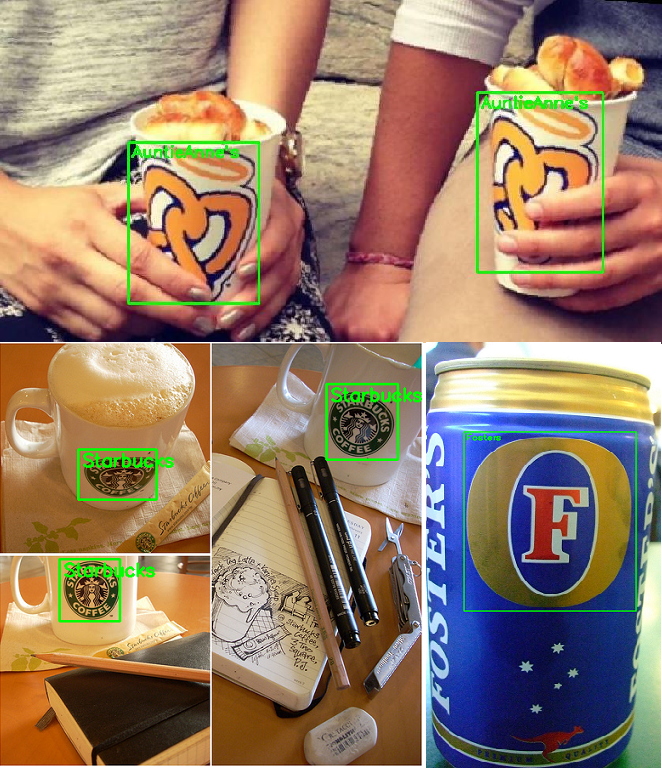} 
\caption{Logo detection through our Two-Layer system} 
\label{fig:full_canvas} 
\end{figure}

For the above reasons, we split the problem into two easier problems. We use an object detection network that
was trained to recognize the bounding boxes of an object class ``logo'' and then send the logo proposals
to a deep image-similarity network that computes feature vectors for the logo candidates allowing to search a prototype database. 
As a result, we can quickly train
both networks with limited amounts of training data, skirting both the costs and
logistical hurdles of collecting annotations, and avoiding the problems
associated with training an end-to-end object detection model with many output
classes. The final output is the same as with traditional object detection
systems; namely, our model returns bounding boxes and class labels.
Fig.~\ref{fig:full_canvas} shows example outputs from our system.

\begin{figure*}[!htbp]
\centering
\includegraphics[width=0.9\textwidth,keepaspectratio]{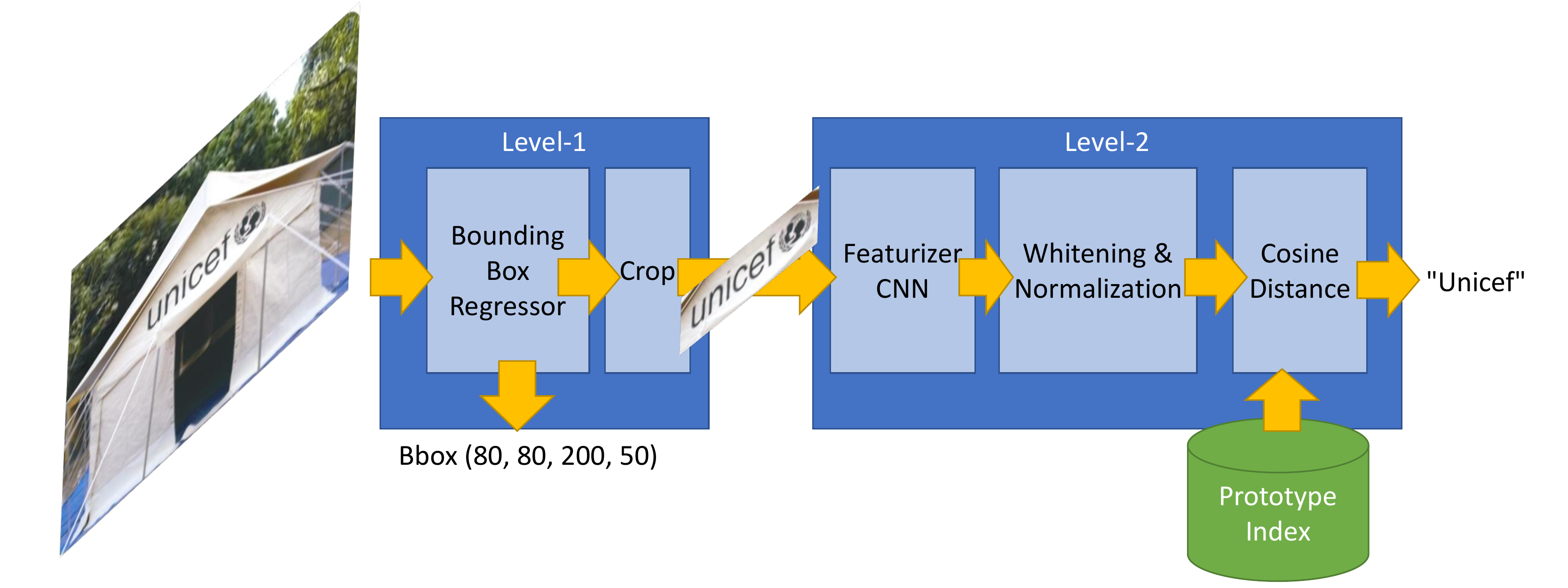} 
\caption
{High-level view of the two-level approach}
\label{fig:architecture}
\end{figure*}

We demonstrate the quality, efficiency, and scalability of our approach
principally on publicly available company logo datasets. First, we show that
our two-layer approach can achieve results that are commensurate with
state-of-the-art end-to-end trained object detection models on small datasets.
We then show how these models quickly deteriorate in quality and practical
ability to train as the number of classes expands. We show that our approach
does not suffer from the same types of growing pains and, importantly, that the
scaling up in ``intelligence'' of the model (in terms of its object knowledge)
does not require a corresponding growth in training data.
Finally, the approach allows to easily update the classifier by simply adding/removing images from the prototype database.
Especially in today's economy, when companies are consistently looking to differentiate 
themselves by (\textit{inter alia}) constantly changing the appearance and
modernity of their brand logo, having a model that does not need to be
consistently re-trained to stay fresh is of the utmost importance.

\begin{figure*} 
\centering
\includegraphics[scale=0.6,keepaspectratio]{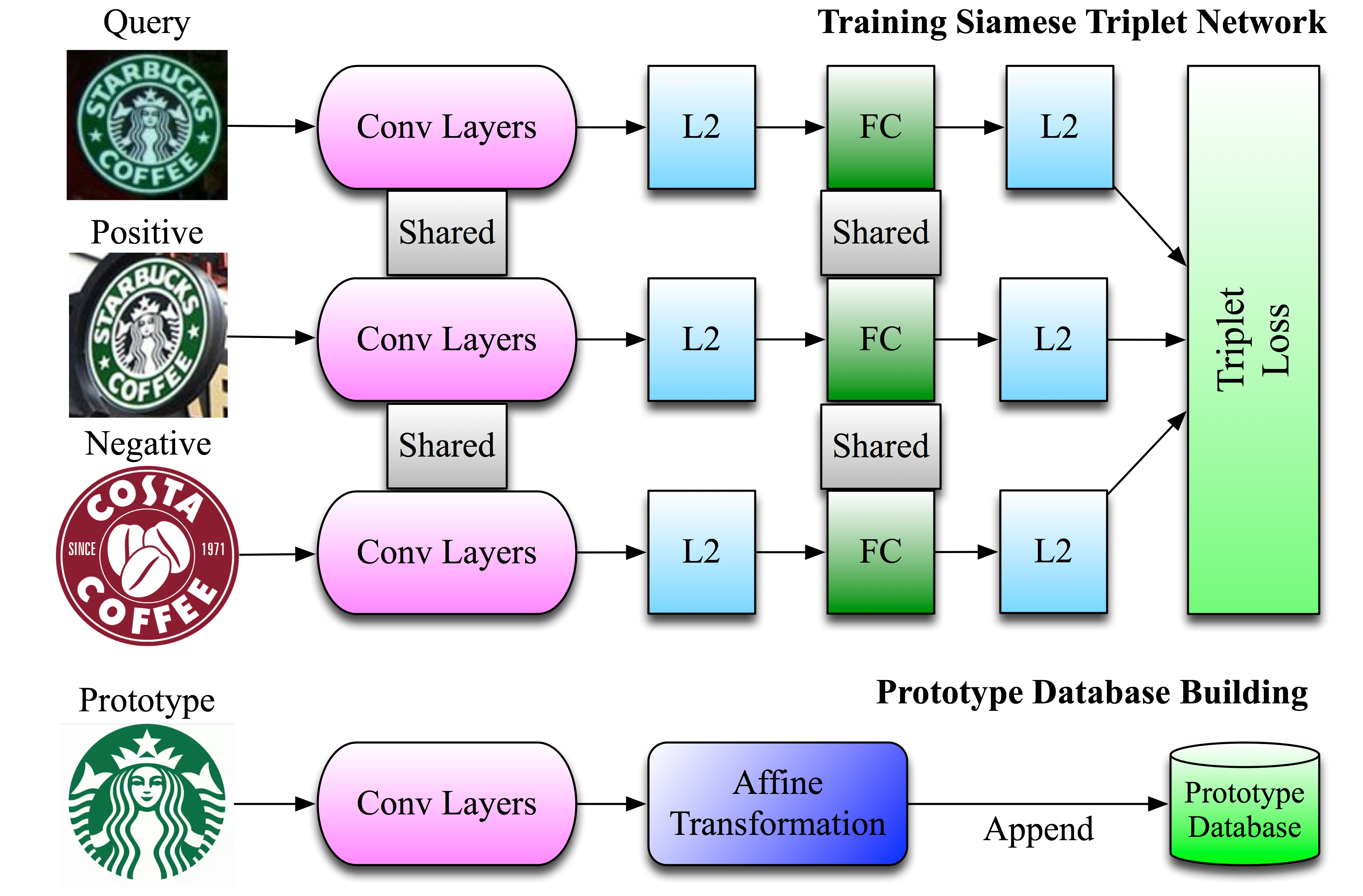} 
\caption
{Training and prototype database generation pipeline} \small Training network
takes query, positive, negative images as input and optimizes for triplet loss.
For prototype database building, we take the last convnet layer and run the
affine transformation to derive the descriptor. 
\label{fig:Training}
\end{figure*}

\section{Related Work}

This section presents related work in the areas of object detection and instance
retrieval.

\textbf{Object detection.} Our work is based in part on techniques for generic 
object detection, either based on a combination of region proposal networks + 
chained convolutional networks \cite{ren2015faster} or end-to-end approaches that
combine a regression target for box coordinates and a classification target for 
identifying classes  \cite{redmon2016yolo9000,liu2016ssd}. The classification 
approach used in these algorithms works well if the number of classes is small. 
However, using softmax for the classification problem does not scale very well 
due to vanishing gradients and increased memory and runtime complexity. At some 
point it requires the use of either approximate versions of softmax, e.g. 
through sampling \cite{JeanCMB14} or by exploiting an underlying
structure in the classes to be recognized 
\cite{Morin05hierarchicalprobabilistic,redmon2016yolo9000}. Due to limits in GPU memory, 
these approaches are still limiting, and we have not yet seen a system that scales 
beyond 10000 classes. Hierarchical approaches also require the explicit 
modelling of those dependencies, which might not be easily apparent in a task 
like logo recognition. Overall, classification-based approaches also suffer
from the need of a large amount of training data per class and the need to
retrain the whole model when changes in the set of classes occur.


\textbf{Instance retrieval.} An alternative to classification is the use of
retrieval-based approaches. In this setup, a high-dimensional feature space is
constructed that prototypical images of a certain class are mapped into.
Classifying a new image then becomes a task of finding the nearest neighbors of
the prototypes in the index, e.g. through cosine similarity. This task predates
deep learning-based approaches, which required the careful design of the
feature space and pre-processing of the query vector, e.g.
\cite{Nister:2006:SRV:1153171.1153548,Philbin07,mikulik_ijcv13,Chum-CVPR-2011}.
Early methods that used image recognition based on convolutional networks
\cite{krizhevsky2012_alexnet,simonyan2014_vggnet,he2015_resnet}  trained on
ImageNet \cite{Russakovsky_ILSVRC15} did not give state of the art results
\cite{razavian2014}. Further refinements to network architectures aimed at rendering them
more invariant to changes such as scaling, lighting, cropping, and
image clutter followed
\cite{babenko2015,KalantidisMO15,GongWGL14,Perronnin_2015_CVPR,ArandjelovicGTP15}.
Fine-tuning networks on top of models pre-trained on the large ImageNet corpus
can increase model robustness further \cite{babenko2015,tuezkoe2017}. A successful approach
for instance retrieval is regional maximum activations of convolutions (R-MAC,
\cite{ToliasSJ15}). It achieves robustness with regards to scaling and
translation by combining several local feature descriptors into a global vector
of fixed length. Another successfull heuristic is to use $l_2$-normalization and PCA whitening \cite{Jegou2012} to normalize and decorrelate the data. PCA whitening can be seen as an additional affine layer that is computed after training where the mean over a particular test set is subtracted and the space is mapped to a decorrelated basis.

There are several approaches to train embeddings using positive and negative training examples.
An early approach coming from the
text domain is DSSM \cite{huang2013_dssm} which compares a source-target
pairing with one or several negative examples obtained by sampling from the
current minibatch. In the image domain, siamese networks \cite{Bromley1993}
combined with a triplet loss \cite{Hoffer2014} are similar.  \cite{GordoARL16}
propose a combination of siamese networks, R-MAC, triplet loss, and techniques
from region proposal networks for achieving state of the art results on
instance retrieval of landmarks.  

For the siamese approaches, it was observed that as the number of classes grows it becomes
important to focus on hard negative examples in the training
process, in a task called hard negative mining. One can distinguish offline
hard negative mining, where a second model is trained to identify hard cases on
the whole dataset \cite{WangSLRWPCW14}, batch hard case mining where batches
are sampled randomly and hard cases are identified within the batch using
exhaustive search \cite{HermansBL17}, and online hard negative mining, where
the method of identifying hard samples is done in a more global fashion and
updated as training proceeds
\cite{ShrivastavaGG16,wang2017_triplet100k,ChenCZH16}. 

\textbf{Combined detection and retrieval approaches.} A case where object
detection is performed with the goal of identifying a large number of instances
is face recognition in images. The area of face detection is a constrained
problem for which solutions exist that can achieve detection results in the
high nineties \cite{viola2004robust}. The task of face recognition can therefore be
split off and take as input the cropped faces from the first step. Similar to
the task presented in this paper, the goal is then to scale up the number of
recognized instances (people). Public datasets exist in the range from 20000 to
8 million instances \cite{guo2016ms,guo2017one}. The task is simpler to ours
since it has to deal with fewer variations in the picture.

\section{Two-Layer Model} At a very high level, our two-layer object detection
model matches the architecture -- or at least the execution flow -- of models
like Faster R-CNN~\cite{ren2015faster} where the first step of object detection
is to predict {\it generic} regions-of-interests (ROI) that are then passed to an
object classification module (see Fig. \ref{fig:architecture}). There is a major problem with scaling up this
paradigm. Region proposal networks are designed to suggest many diversely
sized and oriented regions that could represent interesting areas of an overall
image. It is then up to the object classification module to determine which of
these proposals indeed represents an object of interest, and then to classify
it. The classification module's accuracy quickly breaks down as the number of
distinct objects it must recognize grows. The softmax becomes diffuse and it
becomes increasingly difficult not only to differentiate objects, but to
determine which proposals consist of an object of interest at all.

In the two-layer approach that we advocate, the efficiency and quality required
of an object detection system can be met by both focusing on the equivalent of a region proposal
network in the form of a full ``object class - no object class'' binary object
detection network (a ``logo - no logo'' model in our running example), and
employing a deep, scalable image retrieval system that works with object
prototypes and limited numbers of high-confidence object location proposals. In
the following two sub-sections, we first describe our object region proposal
network (the ``logo - no logo'' model, Level-1), and then our deep, scalable image
retrieval system (Level-2).

A recent paper \cite{tuezkoe2017} discusses a very similar two level approach for logo detection.
The main differences to our paper is that our experiments focus on the scalability of the approach, i.e. 
how it behaves for high number of different classes (logos) to be detected while the former paper focuses
on how the approach behaves if the test set is comprised of a logo set that is disjoint from the training set.
Another difference is that we use a triplet network architecture for retrieval.

\subsection{Level-1: Object Region Proposal Network}

Our object detection framework is posited as two highly distinct steps in a
pipeline. The first step predicts \emph{category-specific} object bounding boxes
where the category is specific to some class of related objects. In our running
example, we focus on brand logos. The key to this paradigm working successfully
is an object detection model that can fully generalize to the high-level object
category in question. Logos might appear at first glance to be in some sense
``easy'' for this task due to their visually similar features. These include
stylized text, common use of concentric and other overlapping primary shapes
(circles, triangles, etc.), and -- perhaps most importantly -- heavy use of some
kind of border around an eye-catching central focus.

While we do not deny that brand logos represent an ideal use case for our
approach (which is clearly one of the reasons we use them as our running
example), all of the noted features are only secondary to the most salient
feature of a logo. That is the logo's context. Logos tend to appear on signs,
which tend to appear on walls or on signposts. Logos also tend to appear on hats
and t-shirts, both of which are highly distinctive to a computer vision system.
Importantly, logos are clearly not the only class of objects that can be
predicted with the help of their surrounding context. Clearly one could apply
this approach to families of objects across a range of specificities. At the
very specific end, we could learn to recognize wine labels. More generically,
one could recognize movie posters, books, etc., all based on their surrounding
context.

As such, it becomes even more apparent why a model like Faster R-CNN is less
applicable as the basis of our Level-1 layer. Instead, we make use of a CNN-based
object detection system from the Single Shot Multibox Detector family of models.
More specifically, for our experiments we use Yolo V2
\cite{redmon2016yolo9000}.\footnote{We discuss exactly how we train our ``logo -
no logo'' model in the Experiments and Results section later in this paper.} The
important distinguishing feature of these networks is that the entire input
image is used at each step of the object detection and so its predictions are
fully informed by the image's global context. 

In some sense our two-layer approach might seem redundant, or even ironic, given
the strengths of the single shot object detection network paradigm. However,
separating the generic \emph{family} of object detection step from the
\emph{specification} of the particular instance of object in that family allows
us to scale up to theoretically an infinite number of family-specific classes.
Because the image retrieval back-end, described just below, is based on image
similarity and thus ranking, we can continuously add classes to the system
without re-training the whole system which would be required with an end-to-end
system like Yolo or SSD. We can scale up without requiring hundreds of new
training examples (in fact we only need a single prototype image), and we do
not suffer from collapsing quality as the number of classes increases. 

\subsection{Level-2: Image Retrieval Back-end} As
discussed throughout this paper, there are inherent issues with scaling up
traditional softmax-based classification models above thousands of image
classes. Additionally, the training data requirements become unmanageable and
adding or removing classes requires an entire re-training of the model. For
these reasons, we eschew multi-class classification and focus on ranking. Our
retrieval model is based on a triplet network architecture. 

Recent work in this area shows learning with a
triplet-based loss has significant improvements over pair-based learning \cite{hoffer2014triplet}. At its
core, triplet learning consists of seeing training samples composed of 3-tuples:
a query ($x_q$), a positive example ($x_p$) and a negative example ($x_n$)
(some literatures refer to the query image as an anchor image). $x_q$ and $x_p$ are
different samples that represent the same class, while the negative sample $x_n$ is
drawn from a different class. \cite{schroff2015facenet} suggests various
methods for picking the negative sample. In training triplet networks,
the feature vectors for $x_q$ and $x_p$ are pushed closer together in the feature space while feature vector vector for $x_n$ is
pushed further away. There are two kinds of losses
for triplet-based networks that are widely used in the literature. These are
margin-based ranking loss~\cite{schroff2015facenet,GordoARL16}, and
ratio-based loss~\cite{wang2014cvpr,wohlhart2015cvpr}.

Our Level-2 retrieval model is a Siamese triplet network as shown in Figure
\ref{fig:Training}. The $d$-dimensional descriptor for an image $x$ is represented by
$f_x$ $\in$ $\mathbb{R}^{d}$. Let us assume that the descriptors for $x_q$, $x_p$, $x_n$
are $f_{x_q} $, $f_{x_p}$ and $f_{x_n}$ respectively. Then, the margin-based
triplet loss is defined as:

\begin{equation} L(x_q,x_p,x_n) = max (0, m + \|f_{x_q}-f_{x_p}\|^2 -
\|f_{x_q}-f_{x_n}\|^2) 
\end{equation}
The margin $m$ is a meta-parameter that has to be chosen empirically. 
During training it controls which training-triplets still contribute to the gradient.
As soon as 
\begin{equation} 
m + \|f_{x_q}-f_{x_p}\|^2 -\|f_{x_q}-f_{x_n}\|^2 < 0 
\end{equation}
the cost $L$ is zero i.e. constant and the corresponding triplet will not contribute to the gradient.
Therefore for small values of $m$ fewer triplets will contribute than for large $m$.
As in our architecture the feature vectors $f_x$ are $l_2$-normalized we have
\begin{equation}
\|f_{x_q}-f_{x_p}\|^2 -\|f_{x_q}-f_{x_n}\|^2 \in [-4, 4] 
\end{equation}
and hence only $m$ values in the range $\left]-4, 4\right]$ need to be considered. 
Unless otherwise stated we use the value $m = 0.6$.

\subsection{Post Processing Whitening}
Following suggestions from the literature, we considered several techniques for post-processing the output of featurization.
One technique, regional maximum activation of convolutions (R-MAC) \cite{ToliasSJ15}, involves max pooling on subregions of the final CNN-layer and averaging the results. We found that for our particular task this did not improve the model and we ended up using just max pooling over the $(x,y,d)$ vectors output by the second to last layer of the convolutional network to arrive at a feature descriptor of length $d$.

The second step involves normalization. We run it through $l_2$-normalization, then an affine transformation step, then another normalization step for mapping the result to a normalized feature space. 

For the affine transform we consider a learned fully-connected layer. We also experimented with using PCA-whitening instead which actually showed favorable results. 
These experiments used different dataset than we use in this paper so we don't report them in more detail.

\subsection{Prototype database} For building the database, we collect typically one
representative image per class (it could be more if needed). 
Generally, these prototype logos are not cropped from photos but are clean clip-art like representations with neutral background.  Next we run these
images through the trained model and the post-processing whitening step to
obtain the $d$-dimensional descriptor $f_{x_c}$ where $x_c$ is any prototype image.
The database is a table of class-name/prototype-descriptor pairs.

\section{Experiments and Results} 

In the following we aim to verify whether the two-layer approach is in fact a good alternative to a pure classification-based approach in the case of stylized object classes. We need to answer the following questions:

\begin{enumerate}

\item What datasets to use for training and testing. Ideally we should be able to verify the technique without creating a large manually annotated corpus. This calls for techniques around synthesizing images.

\item Which object detection algorithm to use for recognizing a broad class like ``logo'' --- ideally with the property of generalizing well to unseen classes

\item How to design the feature space of the retrieval model.

\item How efficient the models are in practice.

\end{enumerate}

\subsection{Datasets}

In order to cover enough ground for answering the above questions, we will consider three types of datasets, each with their advantages and disadvantages, focusing on the brand detection task.
As our Two-Level approach is retrieval  based we need three kinds of sub-sets for evaluation:
\begin{itemize}
\item A training set to train the logo no-logo detector (Level-1) and the featurizer (Level-2).
\item A test set. 
\item A set of logo prototypes that covers all the classes we want to be able to detect. 
\end{itemize}

\textit{Public sets}: The probably most widely used logo-related dataset is Flickrlogos-32 \cite{Flickr32}. 
This set does not come with logo prototypes. We gathered prototypical logo images for these brands to build a retrieval database. For some of the classes (e.g. Starbucks) we obtained more than one logo representation, e.g. if there are different prototypical depictions used or if the logo has changed over time. 

\textit{Large (internal) set}: We use an internal pair of sets we call MSR1k-Train and -Test. The sets are comprised of 1050 brand classes. The training set consists of about 120.000 photographs, so on average 114 per brand (though there might be more than one logo visible in a photograph). The test set contains around 7000 images, so slightly less than 7 per brand.

\textit{Logo prototype set}: From the combined list of brands from the Flickr sets and the MSR1k set we create a ``logo set'' comprising of just the logo prototypes. We add to this a set of logos obtained through Microsoft's knowledge graph (Satori). We ended up with a set of 9000 brands in total. We gathered prototypical logo images for these brands to build a retrieval database. For some of the classes (e.g. Starbucks) we obtained more than one logo representation, e.g. if there are different prototypical depictions used or if the logo has changed over time.

\begin{figure} \begin{center}\includegraphics[width=1.0\columnwidth,keepaspectratio]{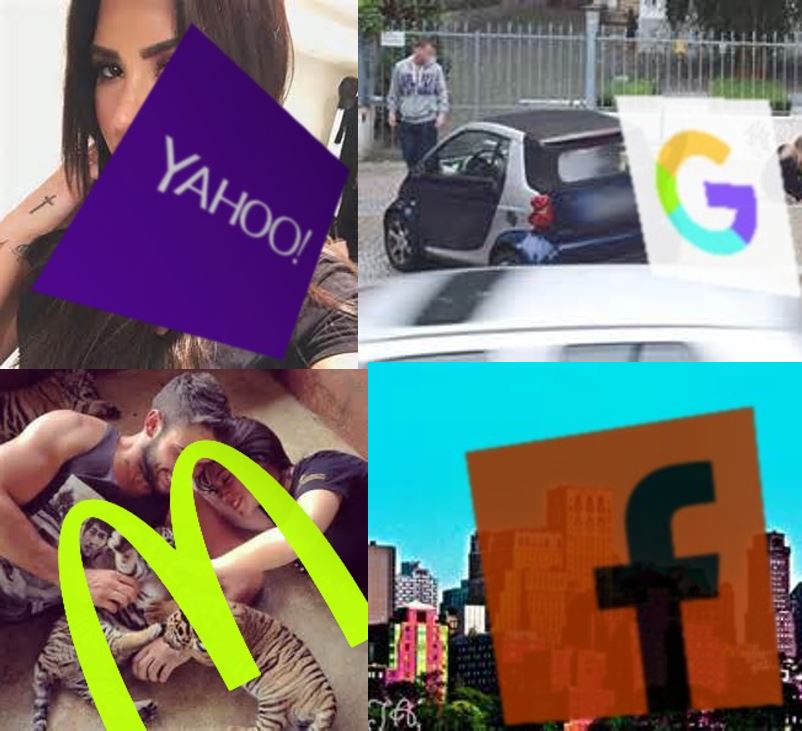} \caption
{Examples of our synthesized images demonstrating applications of projections, rotation, transparency, and color changes.}\label{figure:synthetic}\end{center}
\end{figure}

\textit{Synthetic set}: In order to show the scalability of the method, we created a synthesized set of  {\it 'in-the-wild'} photographs that comprises these 9000 different types of logos. We use in total 9 million photos of natural surroundings, obtained from bing's image search index. Pairs of a logo class and a natural image are randomly selected, and the logo is ``stamped'' into the photo by applying a set of randomly selected transformations:

\begin{itemize}

\item Projection, scaling and rotation
\item Color transformations
\item Scaling of brightness. Inversion of brightness. Adding brightness gradients.
\item Blurring
\item Transparency

\end{itemize}

These transformations create both the transformed images (now containing logos) as well as their bounding boxes which are used as ground truth. Examples for the generated set are shown in Fig.~\ref{figure:synthetic}.

\subsection{Logo Detection}

\begin{table}[!htbp] 
\caption{mAP of Level-1 detection performance trained on MSR1k-Train set (single class: logo)}
\label{table:l1-r1} 
\centering 
{\renewcommand{\arraystretch}{1.4}
\begin{tabular}{ l l } \toprule 
& MSR1k-Test \\ 
\midrule 
FasterRCNN-Resnet-101 & $0.505$  \\ 
Faster RCNN-Inception & $0.525$ \\ 
Yolo V2 & ${\bf 0.623}$  \\ 
\bottomrule \end{tabular} } \end{table}

\begin{table}[!htbp] 
\caption{Generalization: mAP over unseen classes (Trained on MSR1k-Train-700split)}
\label{table:l1-r2} 
\centering 
{\renewcommand{\arraystretch}{1.4}
\begin{tabular}{ l l } \toprule 
& MSR1k-Train-300split \\ 
\midrule 
FasterRCNN-Resnet-101 & $0.441$  \\ 
Fast RCNN-Inception & ${\bf 0.464}$ \\
Yolo V2 & $0.448$ \\ 
\bottomrule \end{tabular} } \end{table}

\begin{figure} \begin{center}\includegraphics[width=1.0\columnwidth,keepaspectratio]{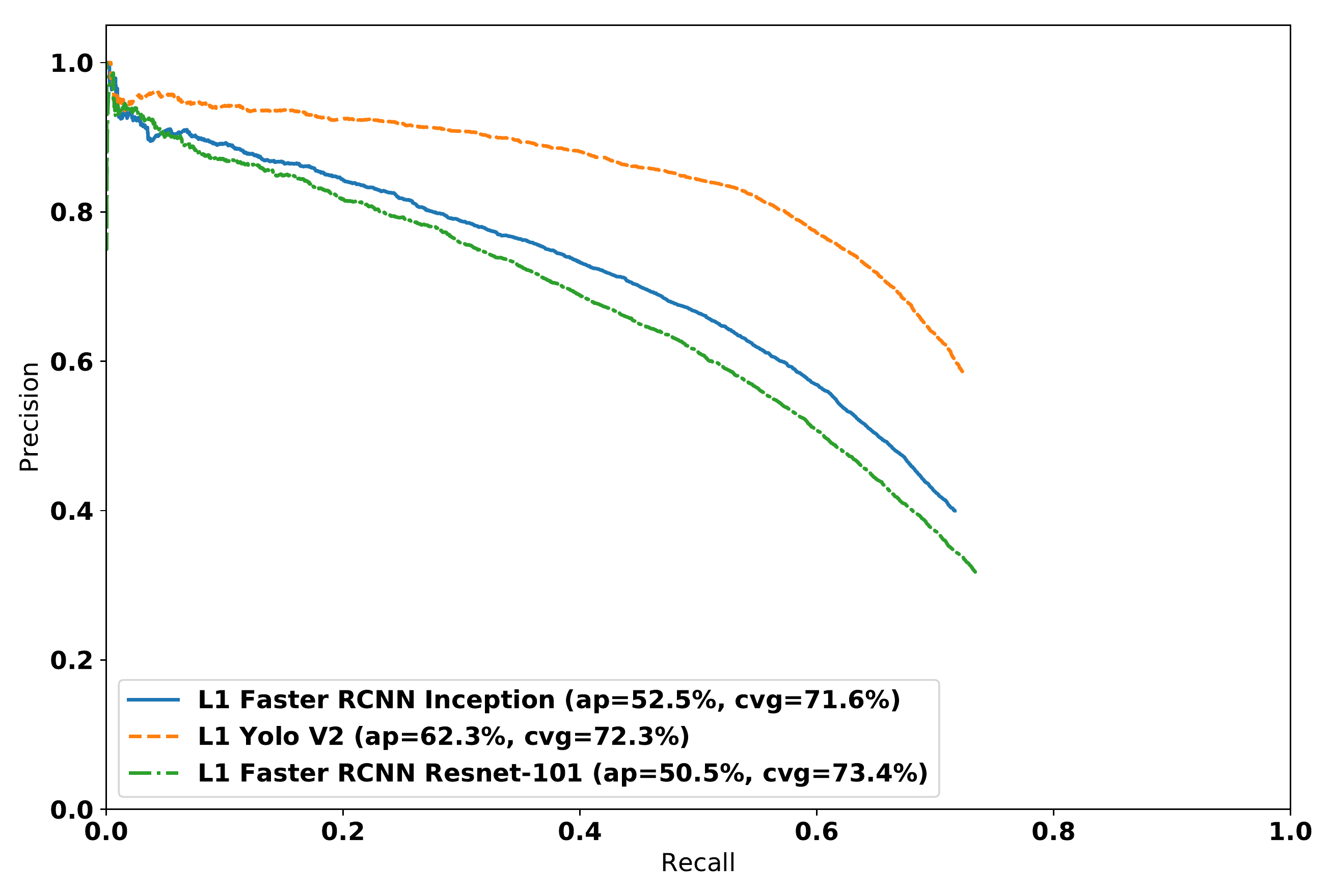} 
\caption {Precision/Recall curves for the logo no-logo detector (i.e. Level-1) on classes seen during training.}\label{fig:l1-detectors}\end{center}
\end{figure}

\begin{figure} \begin{center}\includegraphics[width=1.0\columnwidth,keepaspectratio]{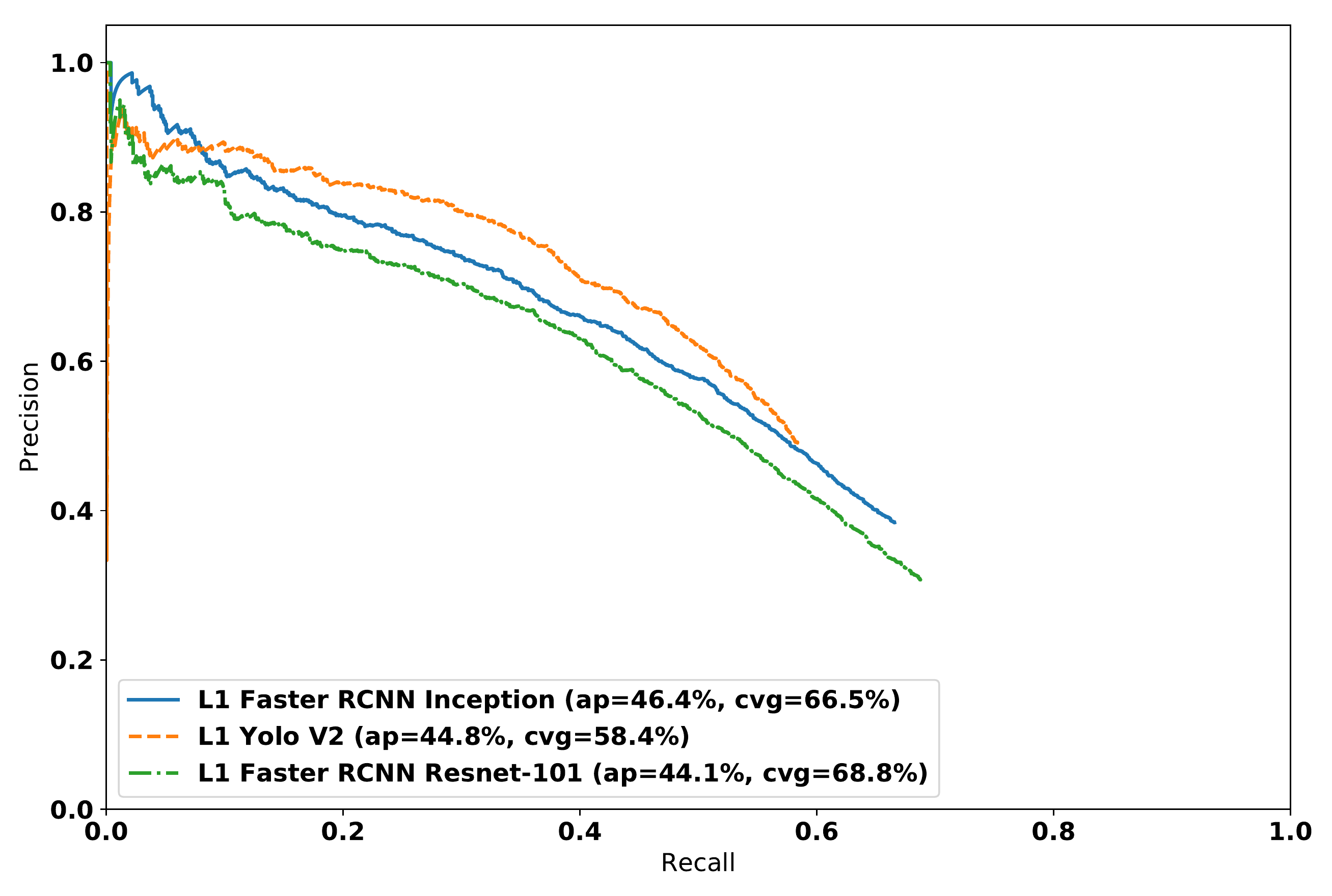} \caption
{Precision/Recall curves for the logo no-logo detector (i.e. Level-1) on classes not seen during training.}\label{fig:l1-unseen}\end{center}
\end{figure}

We aim to choose an object detector that can a) produce good results on logos when they were seen during training and that can b) generalize well to unseen classes. We thought that this detector would ideally take the whole image into account to determine where a particular region should be further inspected, and so we hypothesized that the single shot multibox detector family would be well-suited for the task. We tested this by running implementations of Yolo against the RCNN family of detectors - see results in Table~\ref{table:l1-r2} and the P/R curve in Fig.~\ref{fig:l1-detectors}.  As we can see in Fig. ~\ref{fig:l1-detectors} the average precision in 95-percentile range remains high on Yolo compared to Faster RCNN based model for learned classes. This level is precision is extremely important for a production-grade system.

In a second experiment we wanted to know whether the technique would generalize to unseen classes. For this we split up the MSR1k training set into a set of 720 randomly selected classes for training and 330 classes for testing. The results can be seen in Table~\ref{table:l1-r2} and in detail in the P/R curve in Fig.~\ref{fig:l1-unseen}. 
The coverage of Faster-RCNN based model is high but too many predictions with lesser precision results in bad user experience.

From these experiments we concluded that YoloV2 would obtain the best results in terms of overall quality for detection and generalization. Yolo has the additional benefit of being very efficient at runtime.

\subsection{Retrieval Model}

\begin{table}[!htbp] \caption{Second layer evaluation over the synthetic stamped test set} 
\label{L2alone} 
\centering 
\begin{tabular}{ l l l  } \toprule Model
Config & Recall@1 & Recall@5 \\    \midrule 
Resnet-18, Margin=0.2 & $0.56$ & $0.74$  \\ 
Resnet-18, Margin=0.8 & $0.79$ & $0.91$  \\ 
Resnet-50, Margin=0.8 & $0.78$ & $0.92$  \\
Resnet-50, Margin=0.6 & $0.90$ & $0.97$  \\ 
Resnet-101, Margin=0.8 & $0.85$ & $0.88$  \\
Resnet-101, Margin=0.6 & ${\bf 0.92}$ & ${\bf 0.98}$  \\ 
\bottomrule \end{tabular} \end{table}

We considered a number of variations of the retrieval architecture. We first considered pretrained networks and experimented with feature vectors obtained by maxpooling over different cnn-layers. 
We then settled on a retrieval model based on the architecture shown in Fig. \ref{fig:Training}, which is based on the triplet network architecture described in \cite{GordoARL16}. 
The basic building blocks of this network are three copies of a convolutional neural network that was pretrained on the imageNet task which we fine tune for logo detection. 
These three sub-networks process query, positive and negative training images respectively.
The weights of the second to last convolutions layer is shared across the sub-networks for query, positive, and negative examples.
We tested different pretrained models, namely: ResNet-18/50/101 models.
The layers are then extended by an $l_2$-normalization step followed by a linear fully-connected layer, followed by another $l_2$-normalization step. We train with Adam~\cite{KingmaB14} and use a learning rate of $10^{-5}$ per sample for the highest layers and $10^{-7}$ for the pretrained convolutional network layers which is reduced to $10^{-8}$ after ten epochs and $10^{-9}$ after another five, to avoid unlearing of the pretrained network. 
We adapted the minibatch size to make the result fit in GPU memory.

We had also experimented with average pooling of the last convolutional layer but it turned out to be less effective than max-pooling. 

We found it is important to tune the margin parameter. In table \ref{L2alone} we show some result for different margin parameters and network depths.

We experimented with hard negative mining, but for the number of classes we have used we did not see a notable improvement of the model. 
This might be due to the fact that we searched for hard cases only in a relatively small random subset of all negative cases. Therefore our hard cases might not have been hard enough. 
Instead we pick the negative samples randomly excluding images showing logos of the query class. We trained the model on the synthetic training set with small held out sets for validation and test. In the set we simulated a successful detection of the Level-1 stage. In other words, the input to the training is the crop of the image containing the logos -- so, it is the input logo subject to color and perspective transformations and also application of a part of the photo in the background.
\subsection{Combined Two Layer Approach} 

In order to get end-to-end results, the crops (contents of the bounding boxes) from Level-1 are scaled to a $224 \times 224$ image and passed to the retrieval layer (Level-2). The output of the Level-1 layer is not thresholded on the score and thus is optimized for recall. Level-2 generates predictions over all (or the top $n$) bounding boxes output by Level-1. For the retrieval layer, we use a confidence threshold to remove results that might be noise. 

We ran evaluation separately on the Flickrlogos-32, MSR1k-Test, and the synthetic sets. For the synthetic set we down-sampled several versions to show how the model scales with the number of classes. For every task we used a retrieval database that was specific to the classes used in the test set. The results can be seen in Table \ref{table:twolayer}.

In table \ref{table:twolayer} for the evaluation of the Two Level approach we trained the network only once and used the result of all test sets only adapting the prototype database. For all other models, we retrained every model on every set to make the measurements. For smaller sets, like Flickrlogos-32, the mAP in our approach is almost equal to the Faster RCNN based models. Faster R-CNN becomes very slow as the number of classes is increased: For the 2k set we could evaluate the model in minutes compared to ~5 hours (or 2.5 seconds/image) for Faster-RCNN. Since we have not tuned Faster RCNN in any way, we treat this as anecdotal evidence. We could not train the 9k model due to out of memory issues.

\begin{table}[!htbp] 
\caption{mAP comparison of Two-Layer approach with other SOA approach} 
\label{table:twolayer} 
\centering 
{\renewcommand{\arraystretch}{1.4}
\begin{tabular}{ l l l l l  } \toprule 
& Flickrlogos-32 & MSR1k-Test & Synth9k \\ 
\midrule 
FasterRCNN-Resnet-101 & $0.80$ & $0.58$   & $N.A.$  \\ 
FasterRCNN-Inception & $0.82$  & $0.61$  &   $N.A.$  \\ 
Two-Layer (Ours\footnotemark) & $0.75$ & $0.52$ & $0.58$
\\ \bottomrule \end{tabular} } \end{table}
\footnotetext{Level-1 trained on MSR1k-Train; Level-2 trained on Synth9K-Train.}

\subsection{Implementations}

For Fast(er) RCNN we have used implementations from the Tensorflow Model repository\cite{tf-zoo}. For Yolo we have used the original implementation (Darknet)\cite{darknet}. For the retrieval model we have used Microsoft Cognitive Toolkit (CNTK)\cite{cntk}.

\begin{table}[!htbp] \caption{Performance of the Two-Layer system} \label
{modelperf} \centering \begin{tabular}{ c | c  r  r } \toprule Bounding boxes
& Device & Level-1(ms) & Level-2(ms) \\  \midrule
\multirow{2}{*}{1 bounding box} & CPU & $369.7$  & $406.6$ \\  
& GPU & $38.2$ & $26.2$ \\  
\midrule \multirow{2}{*}{4 bounding boxes} & CPU & $372.8$ & $1626.4$ \\  
& GPU & $38.0$ & $104.8$  \\  \bottomrule \end{tabular}
\end{table}

In Table~\ref{modelperf} we show the performance of
the two layer system at inference time. We show the $95^{th}$ percentile
measurements over the 500 runs. The
runs are repeatedly carried out over same image with 1 bounding box proposal (no caching).
Our experiments were carried out on CPU[Intel(R) Xeon(R) CPU E5 v4 @ 2.10GHz]
and GPU[Nvidia GeForce GTX 1080 Ti]. We compare the performance of each layer on
CPU/GPU. In the numbers for Level-1 and Level-2 we ignore contributions from 
preprocessing of the images, similarity distance computation 
etc. Instead, all of these parts are included in overall run time. 
The overall time of GPU run is slightly higher than that of Level-1/Level-2 as it has also CPU
to GPU and vice versa transfer times.


\section{Discussion} 

In this paper, we have shown how to scale object detection beyond what a softmax-based classifier could do. This is achieved through an approach that avoids the full softmax by doing only pairwise comparisons at training time, and by using a prototype database for similarity computations. We were able to transfer the technique of Siamese networks to the task of logo recognition. The results show that the two-level approach exhibits some nice properties which we discuss below: A simple Level-1 model, a scalable Level-2 model, and adaptability to new classes without retraining. We will also discuss shortcomings and future work.

\subsection{Simple Level-1 Layer}

The Level-1 layer is a pure bounding box regressor and thus quite fast and small. As such it lends itself to being executed on a mobile device in real-time fashion. We are seeing that the model generalizes reasonably well to unseen classes and thus would not have to be retrained for every change in the prototype database. 

\subsection{Scalable Level-2 Layer}

For the retrieval system we showed how to scale to 9000 classes. We think that the concepts can be applied to sets that are several orders of magnitude larger if proper hard negative mining is applied.

\subsection{Dynamic update of the index} It is just impossible to capture every
real world logo and train the model over it. There are millions of logos in the
world and thousands of logos are generated every day. For any production grade
system, the expectation is to not just recognize standard real world logos but any
other new logos that the customer wants to recognize but the current system does not
necessarily capture. One particular example of this scenario is the marketing
campaign of companies. Some campaigns are short-lived but highly targeted to
customers with sets of new logos.  Our current Two-Layer system is capable of
addressing such scenarios. Since the system searches the latent representations
of the logos, it can be extended to detect the additional logos by just adding the latent representations of the new logos.
The core retrieval model does not need to be retrained. Hence the prototypes database can
be extended with new logos and inference part will start recognizing such
new logos populated. The other way around is equally true and in fact it is the
use case for few customers - to detect only few logos of customer's choice with
high recall by using a small database for only those logos of interest.

\subsection{Synthesized training data}

We were able to obtain reasonable results for this task by synthesizing training data. So, the model was forced to learn the set of transformations that were set heuristically (e.g. changes in perspective). This can of course only be a first step, and future work should look for a more general approach to the problem.

\begin{figure} \centering
 \includegraphics[width=1.0\columnwidth]{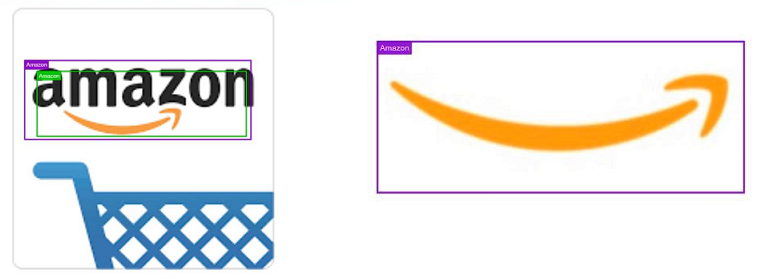} \caption{Amazon logos
through our system} \small On the left side, the Amazon logo is detected by both
systems - particularly on the specific design on which YOLO V2
\cite{redmon2016yolo9000} model has been trained. On the right side, our
Two-Layer system detects the new logo and YOLO V2 \cite{redmon2016yolo9000}
model fails to detect new logo.  \label{fig:newAmazon} \end{figure}

\subsection{Future Work} 

The model that we used certainly requires either more data or a better architecture to capture the underlying distributions more precisely. On the one hand, the model is to some extent resilient to
logo changes. We found that the approach works well with occluded logos, different lighting conditions, background textures etc. Anecdotally we found e.g. that the Amazon logo was recognized by looking only at the arrow, although the prototype database only contained the logo with the amazon text (see Fig. \ref{fig:newAmazon}). However, it is clear that the model is not precise enough -- and has not seen enough data -- for capturing the underlying distributions of logo types in an exact manner. For example, it might accept a logo for ``VOSS'' as being similar to ``BOSS''. Solving this problem might require more specific models for the logo detection task which we have not explored in this paper. The logo detection task is in many ways similar to OCR, since many logos can be broken down into geometric primitives like characters.

The clearest limitation of our approach is that it can only work very well for
families of objects that share some generic characteristics. For example, the Level-1
can do a good job at recognizing logos because they are both similar in some
ways (geometric forms, stylized text, etc.) and show up in similar contexts (a
logo will often show up inside a rectangle on a sign or on a t-shirt). 

%

\bibliographystyle{plain} \bibliography{logo}

\end{document}